\documentclass[twocolumn]{HTL-author}
\DOI{2012.0357}
\begin{document}

\title{Real-time Geometry-Aware Augmented Reality in Minimally Invasive Surgery}

\author{Long Chen$^{1}$\thanks{http://longchen.uk}, Wen Tang$^{1}$ and Nigel W. John$^{2}$}

\address{$^{1}$Bournemouth University\\
$^{2}$University of Chester}

\historydate{To appear in Healthcare Technology Letters}

\abstract{The potential of Augmented Reality (AR) technology to assist minimally invasive surgeries (MIS) lies in its computational performance and accuracy in dealing with challenging MIS scenes. Even with the latest hardware and software technologies, achieving both real-time and accurate augmented information overlay in MIS is still a formidable task. In this paper, we present a novel real-time AR framework for MIS that achieves interactive geometric aware augmented reality in endoscopic surgery with stereo views. Our framework tracks the movement of the endoscopic camera and simultaneously reconstructs a dense geometric mesh of the MIS scene. The movement of the camera is predicted by minimising the re-projection error to achieve a fast tracking performance, while the 3D mesh is incrementally built by a dense zero mean normalised cross correlation stereo matching method to improve the accuracy of the surface reconstruction. Our proposed system does not require any prior template or pre-operative scan and can infer the geometric information intra-operatively in real-time. With the geometric information available, our proposed AR framework is able to interactively add annotations, localisation of tumours and vessels, and measurement labelling with greater precision and accuracy compared with the state of the art approaches.}

\maketitle

\section{Introduction}
Laparoscopic surgery is a Minimally Invasive Surgical (MIS) procedure using endoscopes with small incisions to carry out internal operations on patients. While MIS offers considerable advantages over open surgeries, it also imposes big challenges on surgeons' performance due to the well known MIS issues associated with the field of view (FOV), hand-eye dis-alignment and disorientation. Augmented Reality(AR) technology can help overcome the limitations by overlaying additional information with the real scene through augmentation of target surgical locations, annotations \cite{Kim2012}, labels \cite{Su2009}, tumour  measurement \cite{Bourdel2017} or even 3D reconstruction of anatomic structures \cite{HaouchineDequidtPeterlikEtAl2013} \cite{HaouchineCotinPeterlikEtAl2015}.    

Despite recent advances in powerful miniaturised AR hardware devices and improvements on vision based software algorithms, many issues in medical AR remain unsolved. Especially, the dramatic changes in tissue surface illumination and tissue deformation as well as the rapid movements of the endoscope during the insertion and the extrusion all give rise to a set of unique challenges that call for innovative approaches. 

 As with in any technological assisted medical procedures, accuracy of AR in MIS is paramount. With the use of traditional 2D feature based tracking algorithms such as those used in \cite{Du2015} ,\cite{Plantefeve2016}, \cite{Kim2012} and \cite{MountneyYang2008}, the rapid endoscope movement can easily cause feature points extracted from the vision algorithms to fall out of the field of view, resulting in poor quality visual guidance. The latest visual SLAM (Simultaneous Location and Mapping) based approaches have the potential to overcome this issue by building an entire 3D map of the internal cavity of the MIS environment, but SLAM algorithms are often not robust enough when dealing with tissue deformations and scene illuminations \cite{Turan2017} \cite{Klein2007} \cite{Klein2007} \cite{Mur-ArtalMontielTardos2015} \cite{NaderMahmoud2016}. Furthermore, in order to meet the demand of high computational performance, sparse landmark points are often used in MIS AR, and augmented information are mapped using planar detection algorithms such as Random Sample Consensus (RANSAC) \cite{Lin2013} \cite{GrasaBernalCasadoEtAl2014}. As a result, AR content is mapped onto planes rather than curved organ surfaces.  

In this paper, we introduce a novel MIS AR framework that allows accurate overlay of augmented information onto curved surfaces, i.e. accurate annotation and labelling, and tumour measurements along the curved soft tissue surfaces. There are  two key features of our proposed framework: 1) real-time computation of robust 3D tracking through a robust feature-based SLAM approach; 2) an interactive geometry-aware AR environment through incrementally building a geometric mesh via zero mean normalised cross correlation (ZNCC) stereo matching.

\section{Related Work}

The traditional MIS AR approaches usually employ feature points tracking methods for information overlay. Feature based 2D tracking methods such as Kanade-Lucas Tomasi (KLT) features \cite{Du2015} \cite{Plantefeve2016}, Scale-Invariant Feature Transform (SIFT) \cite{Kim2012}, Speeded Up Robust Features (SURF) \cite{Kumar2014}, even those methods specifically designed to cater for the scale, rotation and brightness of soft tissue \cite{MountneyYang2008} have a major drawback for AR, because selected feature points extracted from vision algorithms must be within the field of view. Therefore, traditional feature tracking methods can severely affect the precision of procedure guidance, especially in surgical scenes where the accuracy is paramount.

Recently, SLAM (Simultaneous localization and Mapping) algorithms have led to new approaches to endoscopic camera tracking in MIS. Originally designed for robot navigation in unknown environments, the algorithms can be adapted for tracking the pose of endoscopic cameras while simultaneously building landmark maps inside the patient body during MIS procedures. SLAM-enabled AR systems not only improve the usability of AR in MIS due to no optical or magnetic tracking devices to obstruct the surgeons' view,  but they also offer greater accuracy and robustness compared with traditional feature-based AR systems. 

Direct-based SLAM algorithms compare pixels \cite{Engel2014} or reconstructed models \cite{Chang2014}  \cite{Turan2017} of two images to estimate camera poses and reconstruct a dense 3D map by minimising the photometric errors. However, direct methods are more likely to fail when dealing with deformable scenes or when the illumination of the scene is inconsistent. Feature-based SLAM systems \cite{Klein2007} \cite{Mur-ArtalMontielTardos2015} only compare a set of sparse feature points that are extracted from images. These methods estimate camera poses by minimising the re-projection error of the feature points. Therefore, feature-based SLAM methods are more suitable for MIS scenes due to it's tolerance to illumination changes and small deformations. Feature-based SLAM such as EKF-SLAM has been used with laparoscopic image sequences \cite{Mountney2006} \cite{Mountney2009} \cite{GrasaBernalCasadoEtAl2014} and a further motion compensation model \cite{MountneyYang2010} and stereo semi-dense reconstruction method \cite{Totz2011} were integrated into the EKF-SLAM framework to deal with periodic deformation. However, the accuracy of EKF-SLAM tracking is not guaranteed and prone to inconsistent estimation and drifting due to the linearization of motion and sensor models approximated by a first-order Taylor series expansion.

The first keyframe-based SLAM -- PTAM (Parallel Tracking and Mapping) \cite{Klein2007} was a breakthrough in visual SLAM and has been used in MIS for stereoscope tracking \cite{Lin2013}. The extension of PTAM -- ORBSLAM \cite{Mur-ArtalMontielTardos2015} has also been tested on endoscope videos with map point densifying modifications \cite{NaderMahmoud2016}, but the loss of accuracy still exists. Furthermore, since feature-based SLAM systems can only reconstruct maps based on sparse landmark-points that barely describe the detailed 3D structure of the environment, the augmented AR content has to be mapped onto a plan through planar detection algorithms such as Random Sample Consensus (RANSAC) \cite{Lin2013}. Although feature-based SLAM is computationally efficient, different to real-life environments, in MIS scenes, flat surfaces are rare and organs and tissues do have smooth and curved surfaces, hence, resulting in inaccurate AR content registration. One example is the inaccurate labelling and measurement of tumour size without accurate surface fit for information overlay, which can be dangerous and misleading during MIS.

In this paper we present a novel real-time AR framework that provides 3D geometric information for accurate AR content registration and overlay in MIS. We propose a new approach to achieving robust 3D tracking through a feature-based SLAM for real-time performance and accuracy required for endoscopy camera tracking. To obtain accurate geometric information, we incrementally build a dense 3D point cloud by using zero mean normalised cross correlation (ZNCC) stereo matching. Therefore, our framework handles the challenging situations of rapid endoscopy movements with robust real-time tracking, while providing an interactive geometry-aware AR environment.

\section{Methods}
\begin{figure}[]
\centering
\includegraphics[width=0.4\textwidth]{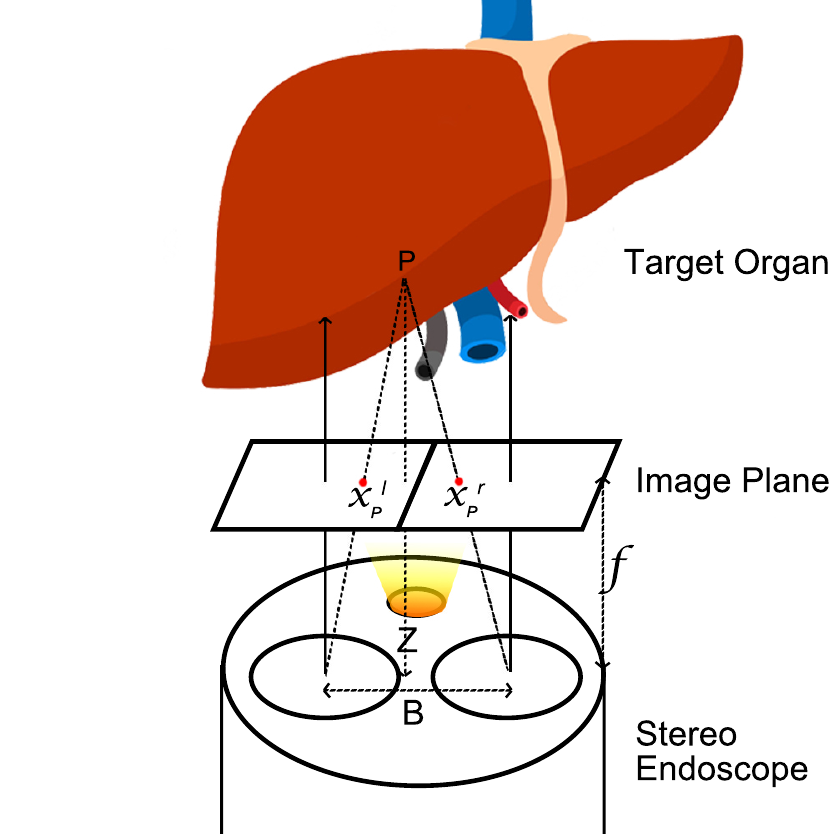}
\caption{By using a stereo endoscope, the 3D position of any point in the view can be directly estimated by using stereo triangulation.}
\label{stereo}       % Give a unique label
\end{figure}
\begin{figure}[]
\centering
\includegraphics[width=0.48\textwidth]{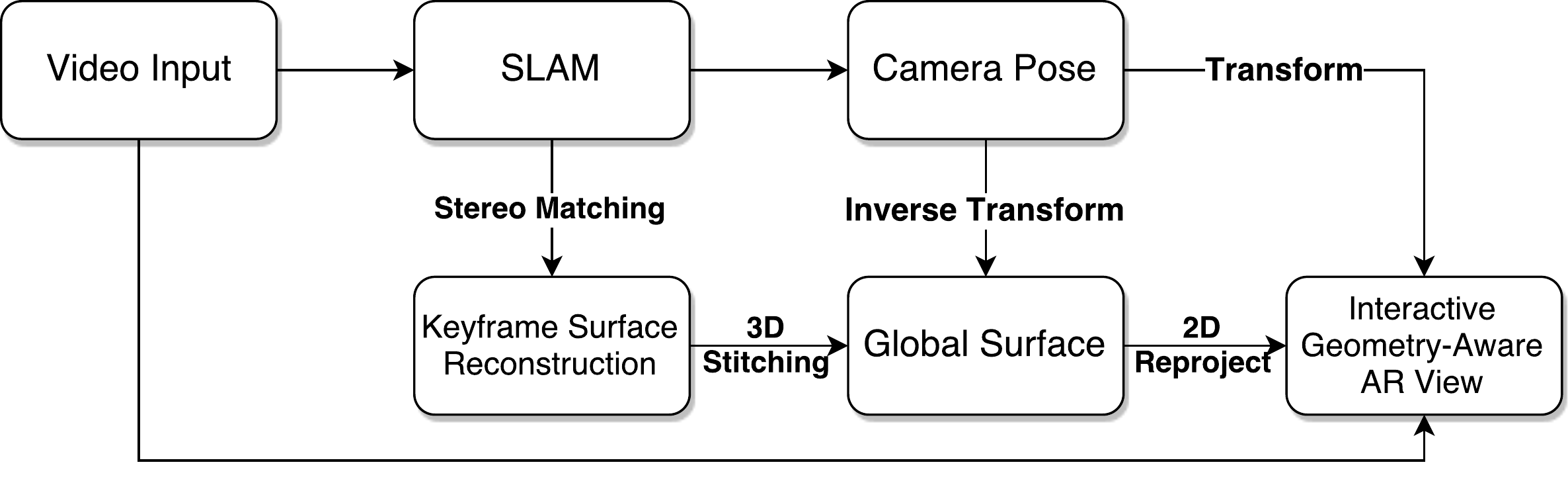}
\caption{Flowchart describing the whole framework}
\label{chart}       % Give a unique label
\end{figure}

As can be seen from the flowchart in Figure \ref{chart}, our proposed framework starts with a SLAM system that can track and estimate the camera pose frame by frame. The following stereo matching algorithm based on ZNCC is used to reconstruct dense surface at each keyframe, which is then transformed and stitched to a global surface based on the inverse transformation of the camera pose. Finally, the global surface is re-projected to 2D based on the camera pose and overlaid on the image frame, serving as an interactive geometric layer. The geometric layer enables the interactive AR applications such as online measurement which will be explained in Section 4.
\subsection{Landmark Point Detection and Triangulation}

In medical interventions, real-time performance and accuracy are both critical. We adopt Oriented FAST and Rotated BRIEF (ORB) \cite{RubleeRabaudKonoligeEtAl2011} feature descriptors for feature points extraction, encoding and comparison to match landmark points in left and right stereo images. ORB is a binary feature point descriptor that is an order of magnitude faster than SURF \cite{BayTuytelaarsVanGool2006}, more than two orders faster than SIFT \cite{Lowe2004} and also offers better accuracy than SURF and SIFT \cite{RubleeRabaudKonoligeEtAl2011}. In addition, ORB features are invariant to rotation, illumination and scale, hence, capable of dealing with challenge endoscope camera scenes (rapid rotating, zooming and changing of brightness).

We apply the ORB detector and find the matched keypoints on left and right images. Let $x_{P}^{l}$ and $x_{P}^{r}$ be the $x$ coordinates on the left and right images, respectively. Assuming the left image and the right image are already rectified, the focal length of both cameras $f$ and the baseline $B$ are known fixed values,  by similar triangles, the depth or the perpendicular distance $Z$ between the points and the endoscope can be found according to similar triangles (see Figure \ref{stereo}).

\begin{equation}
\label{equ_triangulation}
\frac{B-\left ( x_{P}^{l}-x_{P}^{r} \right )}{Z-f}=\frac{B}{Z}\;\;\; \Rightarrow \;\;\; Z=\frac{f\cdot B}{ d_{P}}
\end{equation}

Where $x_{P}^{l}-x_{P}^{r}$ is the disparity $d_{P}$ of the two corresponding keypoints in the left and the right images detected by the ORB feature.

We then perform a specular reflection detection by removing the keypoints that have intensities above a threshold for efficiency. This could effectively remove the influence of specular reflections from the next stage of computation.

\subsection{Frame by Frame Camera Pose Estimation}

Any AR application requires the real-time frame by frame tracking to continuously update the overlay positions. To meet the real-time requirement, after initialization, we employ the constant velocity motion model used by MonoSLAM \cite{Davison2007} to roughly estimate the position $r_{t+1}$ and quaternion rotation $q_{t+1}$ of the camera position based on the current linear velocity $v_{t}$ and angular velocity $w_{t}$ in a small period $\Delta  t$:

\begin{equation}
\left.\begin{matrix}
r_{t+1}=r_{t}+v_{t}\cdot \Delta  t\\ 
q_{t+1}=q_{t}\times q \left ( \omega_{t} \cdot\Delta t \right ) \\ 
v_{t}=v_{t-1}+a_{t-1}\cdot \Delta t \\ 
\omega_{t}=\omega_{t-1}+\alpha_{t-1}\cdot \Delta t 
\end{matrix}\right\}
\end{equation}

Based on the predicted camera pose $(r_{t+1},q_{t+1})$, the potential regions where the feature points may appear on the image are estimate by re-projection of 3D points, hence reducing searching areas and computational cost.

A RANSAC procedure is then performed to obtain the rotation and translation estimations from the set of all the inlier points. During each RANSAC iteration, 3 pairs of corresponding 3D points from current point set $ p_{t}^{i}$ and point set in next period $ p_{t+1}^{i}$ are selected randomly to calculate the rotation matrix $R$ and the translation $t$, which minimizes the following objective function:

\begin{equation}
min\sum_{i=1}^{n}\left \| p_{t}^{i}-\left ( R \ast p_{t+1}^{i}+T \right ) \right \|
\end{equation}
From the set with smallest re-projection error, the set of outlier points is rejected and all the inliers are used for a refinement of the final rotation and translation estimations.

During the inlier/outlier identification scheme by RANSAC, false matched ORB feature points, moving specular reflection points and deforming points are effectively rejected. This is a very important step for a MIS scene where the tissue deformation caused by respiration and heartbeat, as well as blood, smoke and surgical instruments can have impact on the tracking stability. Therefore, at this stage, we use the strategy to filter out any influence caused by occlusion and deformation to recover the camera pose. Indeed, the deformable surface is an unsolved challenge in MIS AR, we address this issue by reconstructing a dense 3D map through a more efficient stereo matching method (see Section \ref{ZNCC}). 

\subsection{Keyframe-based Bundle Adjustment}

As our camera pose estimation is only based on the last state, the accumulation of error over time would cause system drifting. However, we cannot perform a global optimization for every frame as this will slow down the system over time. We follow the successful approach of PTAM \cite{Klein2007} and ORBSLAM \cite{Mur-ArtalMontielTardos2015} in correcting system drafting, which use the keyframe-based SLAM framework to save ``snapshots'' of some frames as keyframes to enhance the robustness of the tracking whilst not increasing computational load on the system. Each keyframe is selected based on the criteria that the common keypoints of the two keyframes are less than 80\% keypoints but the total number exceeds 50. 

Once a keyframe is assigned, bundle adjustment (BA) is applied to refine the 3D positions of each stored keyframe $KF_{i}$ and the landmark points $P_{j}$ by minimising the total Huber robust cost function w.r.t the re-projection error between 2D matched keypoints $p_{i,j}$ and camera perspective projections of the 3D positions of keyframes $KF_{i}$ and the landmark points $P_{j}$:

\begin{equation}
\underset{KF_{i},P_{j}}{\arg\min}\sum_{i,j}\rho_{h} \left ( \left \| p_{i,j}-CamProj\left (KF_{i},P_{j} \right ) \right \| \right )
\end{equation}

\subsection{ZNCC Dense Stereo Matching}
\label{ZNCC}

We create a feature-based visual odometry system for the endoscopic camera tracking and landmark points mapping, which takes into account of illumination changes, specular reflections and tissue deformations in MIS scenes. However, as the sparse landmark points can barely describe the challenging environment of MIS scenes, we perform a dense stereo matching upon the landmark points to create a dense reconstruction result. 

The dissimilarity measure used during the stereo matching is a patch-based ZNCC method. The cost value $C(p, d)$ for a pixel $p$ at disparity $d$ is derived by measuring the ZNCC of the pixel in the left image and the corresponding the pixel $p-d$ in the right image:

\begin{equation}
\resizebox{.9\hsize}{!}{$C\left ( p,d \right )=\frac{\sum_{q\in N_{p}}\left ( I_{L}\left ( q \right )-\bar{I}_{L}\left ( p \right ) \right )\cdot \left ( I_{R}\left ( q-d \right )-\bar{I}_{R}\left ( p-d \right ) \right )}{\sqrt{\sum_{q\in N_{p}}\left ( I_{L}\left ( q \right )-\bar{I}_{L}\left ( p \right ) \right )^{2}\cdot \sum_{q\in N_{p}}\left ( I_{R}\left ( q-d \right )-\bar{I}_{R}\left ( p-d \right ) \right )^{2}}}$}
\end{equation}
where $\bar{I}\left ( p \right ) =\frac{1}{N_{p}}\sum_{q\in N_{p}}I\left ( q \right )$ is the mean intensity of the patch $N_{p}$ centered at $p$. 

ZNCC is proven to be less sensitive to illumination changes and can be parallelised efficiently on a GPU \cite{StoyanovScarzanellaPrattEtAl2010}. A WTA (Winner-Takes-All) strategy is applied to choose the best disparity value for each pixel $p$, followed by a convex optimization to solve the cost volume constructed by Huber-$L^{1}$ variational energy function \cite{ChangStoyanovDavisonEtAl2013} for a smooth disparity map. We used the GPU implement of ZNCC and convex optimization for the efficient disparity map estimation and filtering in real-time.

\subsection{Incremental Building of Geometric Mesh }
\begin{figure}[]
\centering
\includegraphics[width=0.47\textwidth]{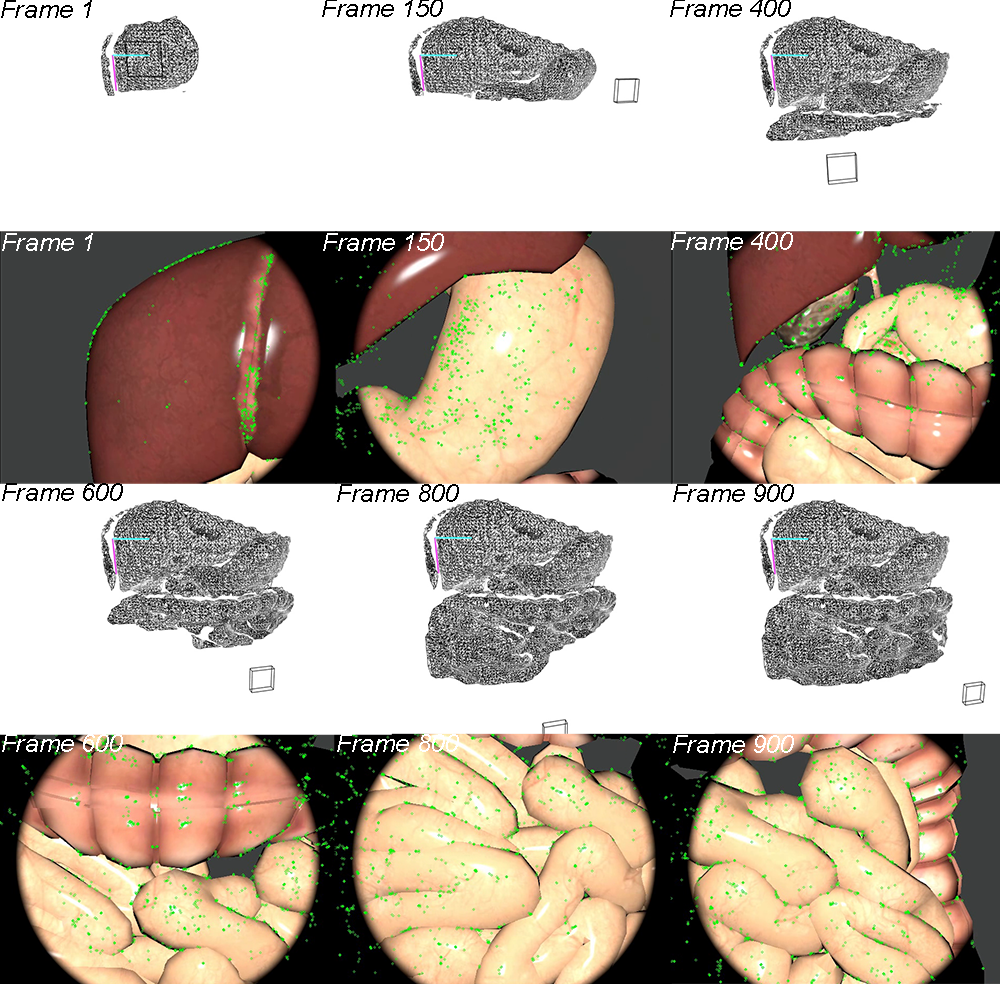}
\caption{Incrementally building the geometric mesh. Rectangular boxes are the estimated camera pose; Green points are detected landmark points.}
\label{increnmental}       % Give a unique label
\end{figure}

The 3D dense points estimated by stereo matching are transformed to the world coordinate system by the transformation matrix from frame space to the world space $T_{f2w}$ that was estimated by our feature-based SLAM system. A fast triangulation method \cite{Marton2009} is then used to incrementally reconstruct the dense points into a surface mesh. Fig. \ref{increnmental} demonstrates the incrementally building process from Frame 1 to 900. The first and third rows are the reconstructed geometric mesh while the second and fourth rows are the current video frames. The geometric mesh can be built incrementally to form a global mesh that can then be re-projected back to the camera's view using the estimated camera pose for the augmented view (see (a) and (c) in Fig. 4 and Fig. 5).

\section{Results and Discussion}
\label{studies}
We have designed a two-parts assessment process to evaluate our AR framework: (i) using a realistic 3D simulated MIS scene as the ground truth study to measure the reconstruction error by measuring the difference between the ground truth values and the reconstructed values; (ii) using a real \textit{in vivo} video acquired from the Hamlyn Centre Laparoscopic/Endoscopic Video Datasets \cite{London2016} \cite{Mountney2010} to assess the quality of applications of our proposed framework i.e. measurements, adding AR labels and areas highlighting.

\subsection{System setup}

Our system is implemented in an Ubuntu 14.04 environment using C/C++. All experiments are conducted on a workstation equipped with Intel Xeon(R) 2.8 GHz quad core CPU, 32G Memory, and one NVIDIA GeForce GTX 970 graphics card. The size of the simulation image sequences and \textit{in vivo} endoscope videos is 840 X 640 pixels. The AR framework and 3D surface reconstruction run in different threads. The 3D surface reconstruction process takes about 200ms to traverse the entire pipeline for each frame. Our proposed AR framework can run in real-time at 26 FPS when the reconstruction only performs at keyframes.

\subsection{Ground truth study using simulation data}
\begin{figure}[]
\centering
\includegraphics[width=0.47\textwidth]{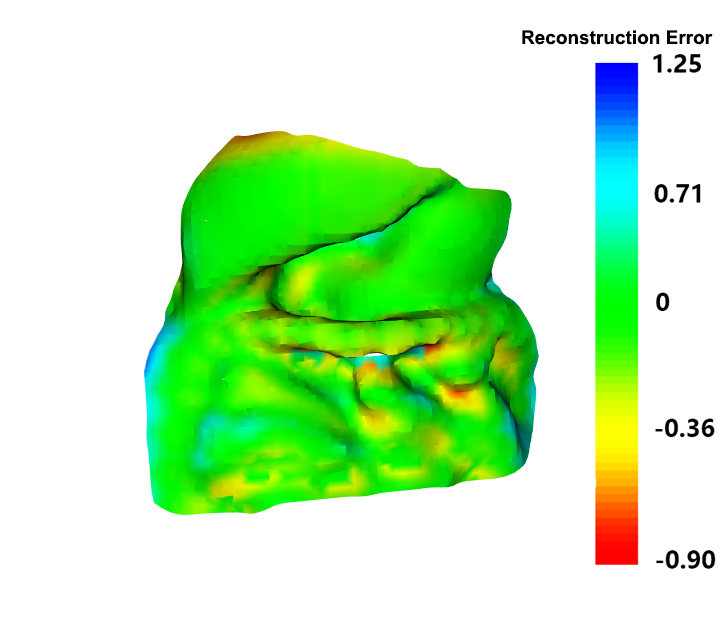}
\caption{Reconstruction error map.}
\label{result}       % Give a unique label
\end{figure}

The performance of our proposed framework is measured in terms of reconstruction accuracy by comparing the reconstructed surface with the 3D model used to render the simulation video.

To quantitatively evaluate the performance of the progressive reconstruction result, we used Blender \cite{Blender2016} -- an open source 3D software to render realistic image sequences of a simulated abdominal cavity scene using a set of pre-defined endoscopic camera movements. The simulated scene contains models scaled to real life size according to an average measured liver diameter of 14.0 cm \cite{Kratzer2003}, and the digestive system is rendered with appropriate textures to make the scene as realistic as possible. The material property is set with a strong specularity component to simulate the smooth and reflective liver surface tissue. The luminance is intentionally set high to simulate an endoscope camera as shown in Fig. \ref{measure} with a realistic endoscopic lighting condition by using a spot light attached to the main camera. We have designed a camera trajectory that hovers around the 3D models. There are a total of 900 frames of image sequences at a frame-rate of 30 fps being rendered, which is equivalent to a 30 seconds video.

Root Mean Square Distance (RMSD) is used to evaluate the overall distance between the simulated and the reconstructed surfaces. By aligning the surfaces to the real world coordinate system, we apply a grid sample to get a series of x,y coordinate points based on the surface area, and then compared the distance of the z value of the two surfaces.

$$RMSD = \sqrt{\frac{1}{mn}\sum_{x=1}^{m}\sum_{y=1}^{n}\left ( Z_{x,y}-z_{x,y} \right )^{2}}$$

The RMSD measurement for the two surface alignments has shown a good surface reconstruction results from our proposed methods, compared to the ground truth surface, the RMSD is 2.37 mm. The reconstruction error map can be viewed in Fig. \ref{result}.

\subsection{Real endoscopic video evaluation}

\begin{figure}[]
\centering
\subfloat[]{\includegraphics[width=0.23\textwidth]{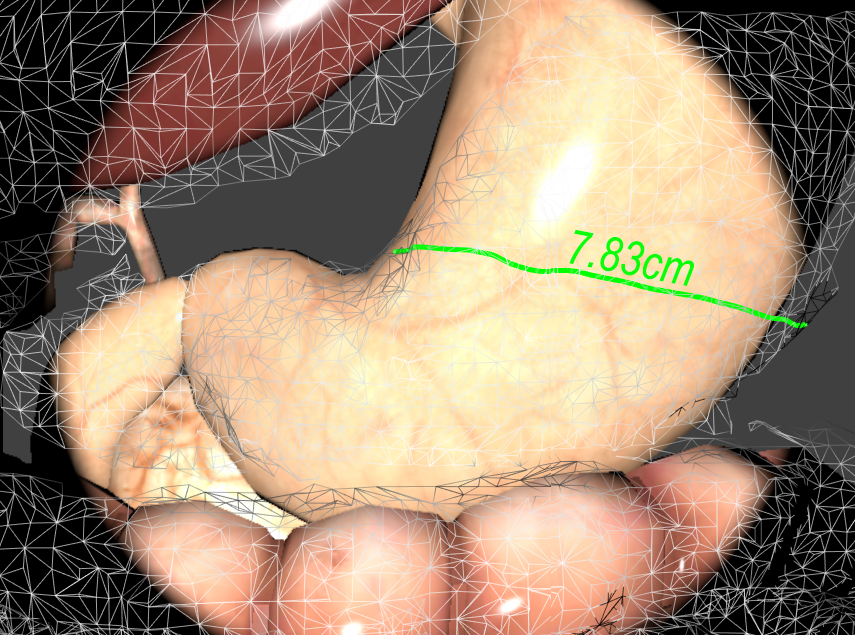}}\
\subfloat[]{\includegraphics[width=0.23\textwidth]{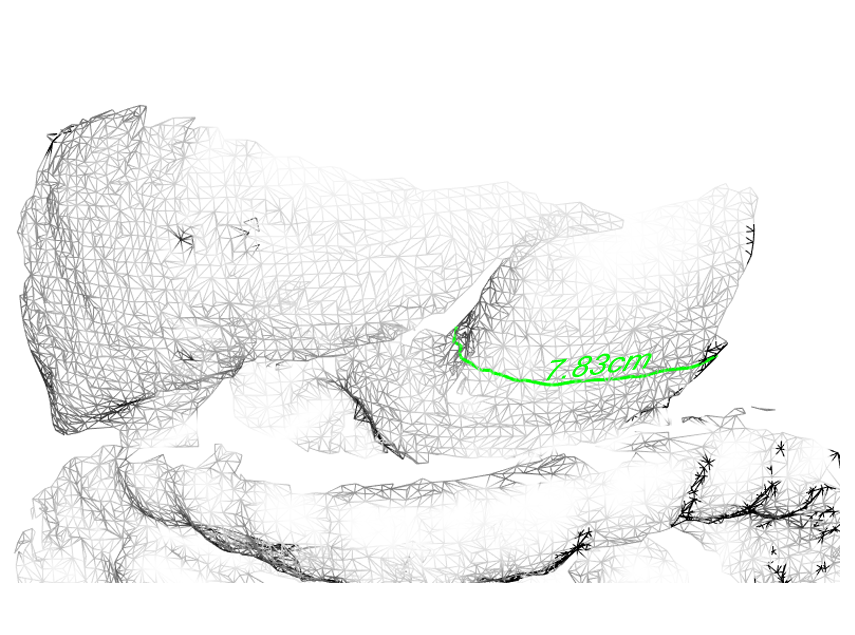}}\\
\subfloat[]{\includegraphics[width=0.23\textwidth]{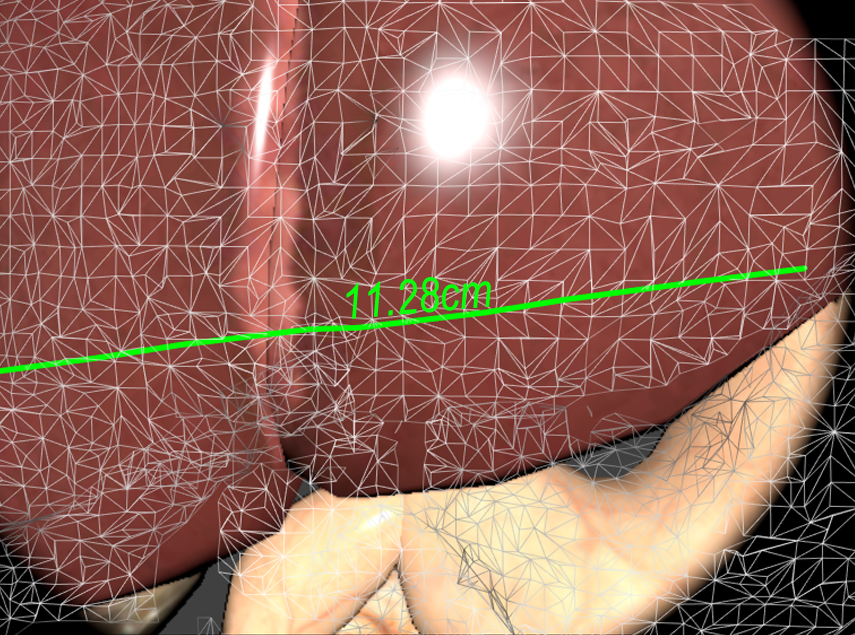}}\
\subfloat[]{\includegraphics[width=0.23\textwidth]{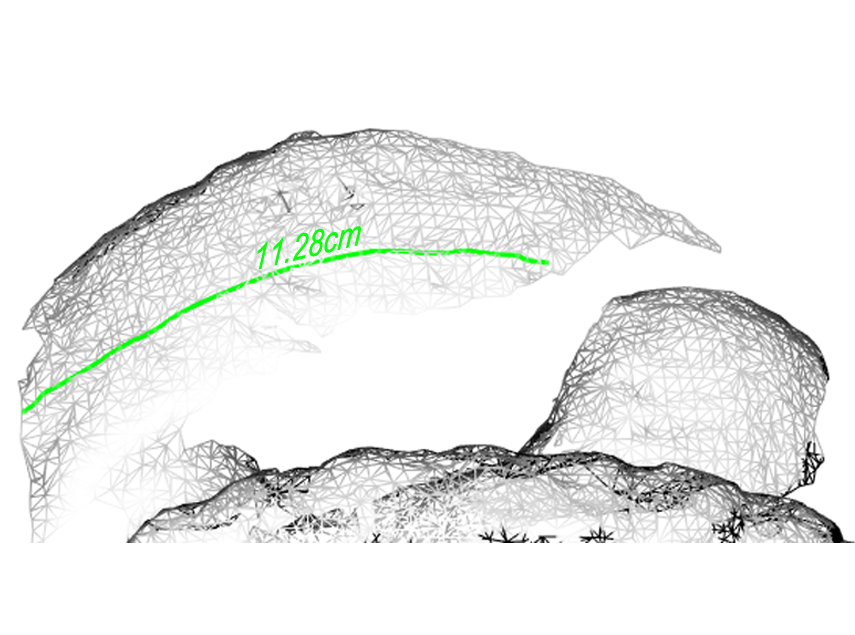}}\\
\caption{Measurement application of our proposed geometry-aware AR framework. Note that the measuring lines (green lines) accurately follow along the curve surface.}
\label{measure}       % Give a unique label
\end{figure}

\begin{figure}[bt!]
\centering
\subfloat[]{\includegraphics[width=0.23\textwidth]{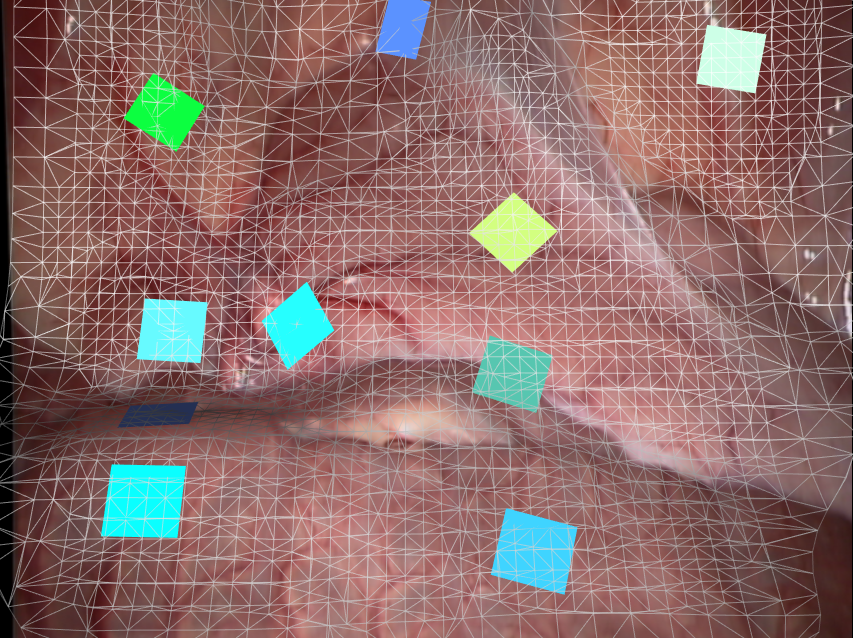}}\
\subfloat[]{\includegraphics[width=0.23\textwidth]{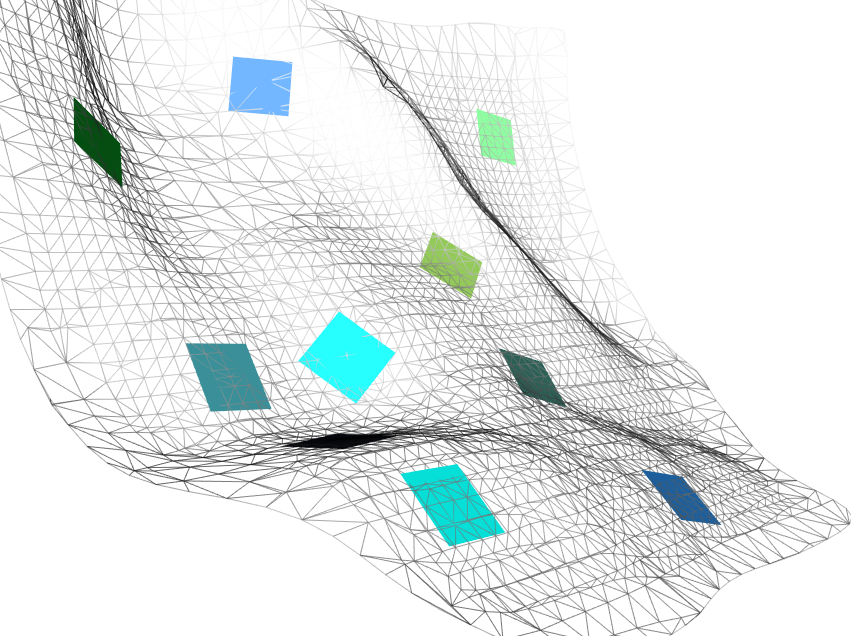}}\\
\subfloat[]{\includegraphics[width=0.23\textwidth]{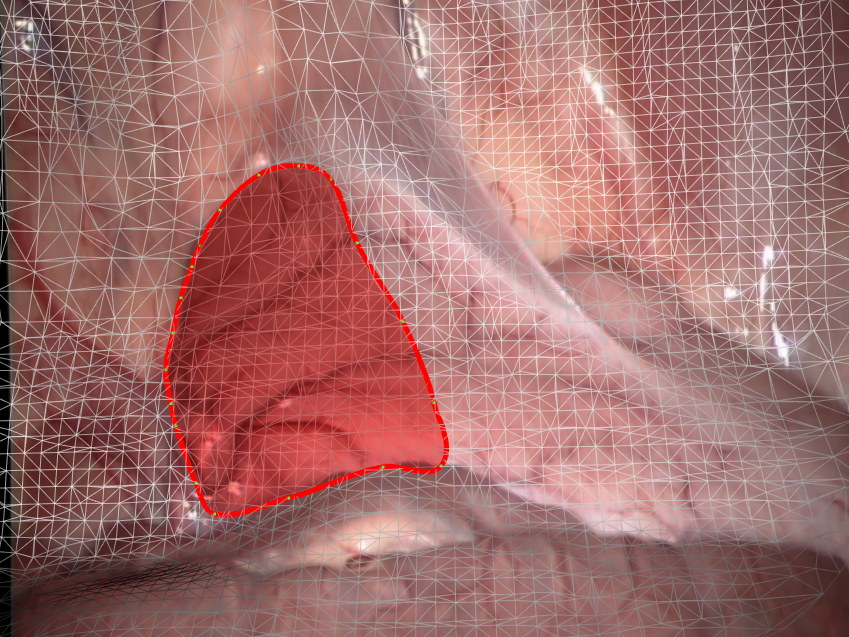}}\
\subfloat[]{\includegraphics[width=0.23\textwidth]{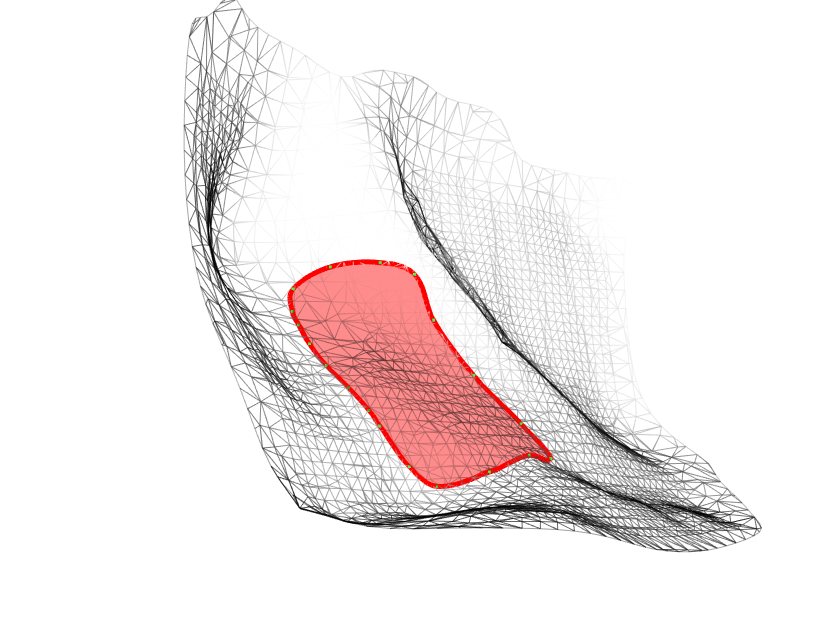}}\\
\caption{Applications of our proposed geometry-aware AR framework: (a) Adding AR labels according to the norm of the geometric surface. (b) The side-view of labels in mesh view. (c) Area highlight and measurement. (b) The side-view of highlighted area in mesh view.}
\label{real}       % Give a unique label
\end{figure}

To qualitatively evaluate the performance of our proposed surface reconstruction framework, we applied the proposed approach on \textit{in vivo} videos that we acquired from Hamlyn Centre Laparoscopic / Endoscopic Video Datasets \cite{London2016} \cite{Mountney2010}. Fig. \ref{real} (a) shows the reconstruction result from our 3D reconstruction framework with the augmented view of \textit{in vivo} video sequences. By clicking the mesh, augmented objects (colored planes) can be superimposed at corresponding positions with correct poses based on the normals of the points at the click locations. Fig. \ref{real} (b) shows the side-view of the mesh; note that the colored planes (which could be labels) are sticking onto the mesh correctly to create a realistic augmented environment. Fig. \ref{real} (c) shows the area highlighting function of our proposed AR framework. And Fig. \ref{real} (d) is the corresponding mesh view. The area highlighting function can be extended to an area measurement and line measurement (such as shown in Fig. \ref{measure}) application once the extrinsic parameters of the camera are known.

\section{Conclusions}
\label{conclusions}
In this paper, we presented a novel AR framework for MIS. Our framework handles the two intertwined issues of tracking the rapid endoscope camera movements and providing accurate information overlay onto the curved surfaces of organs and tissues. By adapting the latest SLAM algorithms, we take a set of innovative approaches at the each stage of the AR process to improve the computational performances and AR registration accuracy. As a result, an interactive real-time geometric aware AR system has been developed. The system is capable of dealing with small soft tissue deformations, rapid endoscope movement and illumination change, which are common challenges in MIS AR. Our proposed system does not require any prior template or pre-operative scan. The system can overlay accurate augmented information such as annotations, labelling, and measurements of a tumour over curved surfaces, greatly improving the quality of AR technology in MIS. 

In future work we will carry out a clinical pilot study. A case scenario will be investigated in collaboration with a practicing surgeon, and comparisons will be made as to the effectiveness of our system with the current procedural approach used.

\section{Conflict of Interest}
None declared.
\bibliographystyle{unsrt}
\bibliography{bibtex}

\end{document}